\documentclass[runningheads]{llncs}
\usepackage[T1]{fontenc}
\usepackage{graphicx}
\usepackage{booktabs}
\usepackage{tabularx} 
\usepackage{amsmath}
\usepackage{float} 
\usepackage{xcolor}
\usepackage{multirow}
\usepackage{makecell}

\usepackage[caption=false,font=footnotesize]{subfig}

\begin{document}
\title{Lake Detection and Water Quality Estimation in Sentinel-2 Data}
\author{Iulia Ple\c su\inst{1}\orcidID{0009-0008-8701-4929} \and
Alexandra B\u aicoianu\inst{1}\orcidID{0000-0002-1264-3404} \and
Ioana Cristina Plajer\inst{1}\orcidID{0000-0002-2666-8215}}
\authorrunning{I. Ple\c su et al.}
\institute{Transilvania University of Bra\c sov, Faculty of Mathematics and Computer Science, Romania\\ %\and
%Springer Heidelberg, Tiergartenstr. 17, 69121 Heidelberg, Germany
\email{iulia.plesu@student.unitbv.ro}, \email{\{a.baicoianu, ioana.plajer\}}@unitbv.ro}
\maketitle              % typeset the header of the contribution
\begin{abstract}
With climate change and increasing human pressure on natural landscapes, inland water resources are becoming progressively scarcer, more vulnerable, and more difficult to manage sustainably. Reliable and automated methods for detecting, monitoring, and assessing surface water bodies are therefore of growing scientific and practical importance. In this paper, we investigate and compare three distinct machine learning architectures for water body identification and monitoring. Their performance is evaluated through quantitative metrics and real-world examples. Furthermore, a direct comparison with classical NDWI thresholding is conducted on a representative test image to highlight differences between data-driven and index-based approaches. This analysis allows us to identify the best-performing model in terms of accuracy, robustness, and practical applicability. Beyond detection, a major challenge for meaningful water quality assessment lies in the consistent and interpretable visualization of spectral water indices. Standard color mapping techniques are often inadequate or potentially misleading for environmental applications. To address this gap, we propose a suite of meaningful color schemes adapted for water quality indices, facilitating clearer interpretation, comparison, and decision-making for human users.

\keywords{Water body segmentation  \and color maps \and quality indices.}
\end{abstract}
\section{Introduction}
Water is a vital resource for life. However, the planet’s hydrological cycle is increasingly affected by climate change and anthropogenic activities \cite{YANG2021115}. Consequently, the growing importance of monitoring water supply sources has led to an increase in research efforts in this field in recent years. This research includes the automatic detection of inland water surfaces, as well as the estimation of their quantitative and qualitative parameters. In this context, satellite imagery, which has become increasingly available for both research and the development of monitoring systems, plays a significant role. 

Compared to coastal waters, accurately detecting inland water bodies is more challenging, particularly due to the presence of other surfaces that may exhibit water-like appearances \cite{kothari2025adversarialrobustnessdeeplearning}. Classical thresholding techniques often fail to achieve accurate segmentation, due to variable weather conditions \cite{CORDEIRO2021112209} and the non-uniform characteristics of the Earth’s surface, such as shadows or dark soil \cite{s21227494}. On the other hand, machine learning (ML)-based approaches face a major challenge related to the limited availability of labeled datasets. Constructing reliable reference datasets over large geographical areas is particularly difficult due to the scarcity of large-scale in situ data collection, as well as the significant financial and time resources required to build and maintain remote sensing databases \cite{CORDEIRO2021112209}.

This paper conducts a comparative analysis of three different deep learning (DL) models for inland water segmentation: a fine-tuned DeepLabV3 as implemented in PyTorch \cite{pytorch_docs} network, a custom U-Net model trained from scratch, and the benchmark DeepWaterMap \cite{article} architecture. The results of the DL segmentation models are further compared to the segmentation using thresholds on classical indices used for water body detection like the Normalized Difference Water Index (NDWI) and the the Modified NDWI (MNDWI) on a specific example of an inland water body, acquired by  Sentinel-2 satellites. 

On top of the performed water segmentation, we use the information provided by the multispectral images of the Sentinel-2 satellite to perform a diagnostic evaluation of water quality indicators, including turbidity and algal biomass. Moreover, we propose a structured visualization framework to mitigate the absence of standardized spectral index color schemes \cite{article2011}, thereby promoting consistency in the qualitative interpretation of environmental information.

\section{Materials and Methods}
In this section we will describe the two models trained for inland water body segmentation and the training procedure, as well as the water quality indices taken into consideration. Furthermore, we describe the color palette selected, for a consistent and unified generation of water quality color maps.

\subsection{Dataset}
For training our deep learning models for inland water body segmentation, we used the optical component of the S1S2-Water dataset \cite{marc_wieland_2024_11278238}, a comprehensive reference dataset designed for the semantic segmentation of surface water bodies. Although the complete dataset includes multimodal triplets consisting of Sentinel-1 SAR data, Sentinel-2 optical imagery, and binary water masks, our analysis focuses exclusively on the Sentinel-2 imagery.
The Sentinel-2 subset includes 65 high-resolution scenes collected from May 2018 to November 2020. These samples span 29 countries and 18 dominant land cover classes, capturing a broad range of topographic settings with varying elevations and slopes.\cite{10321672}. Each scene adheres to the standard Sentinel-2 tiling scheme and spans an area of $100 \times 100$ km. During training, 25 patches were randomly drawn from each scene using a water-biased sampling strategy that required each patch to contain at least 1.5\% water coverage. The data are provided in Level-1C (L1C) format, corresponding to top-of-atmosphere (TOA) reflectance \cite{10321672}. The spatial resolution for each spectral band is of $10 \times 10$ m. 

To further evaluate the qualitative performance of the models, as well as to compare with classical results obtained by thresholding the NDWI and the MNDWI, we downloaded from the Copernicus Data Space Ecosystem (CDSE) platform \cite{cdse_documentation} a scene depicting Lake Dumbrăvița, Brașov, Romania. The multispectral data were acquired on July 19, 2022, under favorable conditions, with a cloud cover of only 6.67\%. The segmentation results obtained on this image are also significant because the scene represents previously unseen data that were not included in the training set. 

\subsection{Model architectures}
This paper evaluates two distinct deep learning architectures for binary water segmentation using multispectral Sentinel-2 imagery: a fine-tuned DeepLabV3 and a custom U-Net implemented from scratch. While the internal structures of these models differ, they were integrated into a unified training framework to ensure a rigorous and fair comparison of their performance on multispectral data. In addition, the performance of the proposed models is evaluated in comparison to the DeepWaterMap architecture, a state-of-the-art convolutional neural network (CNN) that has been specifically tailored for extracting water bodies trained on Landsat 7 imagery \cite{article}.

In contrast to several models proposed in the literature that rely solely on RGB spectral bands \cite{9749269}, our approach uses six of the thirteen spectral bands, namely B02, B03, B04, B08, B11, B12, available in Sentinel-2 imagery as input to both models.  This particular group of bands was chosen to supply the spectral data required for computing the hydrological indices and environmental quality measures described in the following sections of this study. The inclusion of these additional spectral bands provides richer information and can help mitigate errors caused by color similarities between water bodies and other dark surfaces on the ground. The description of the selected bands and their central wavelengths can be found in the specific literature \cite{pour2023geospatial}.\\

\begin{figure}[h]
    \centering
    \includegraphics[width=0.76\textwidth]{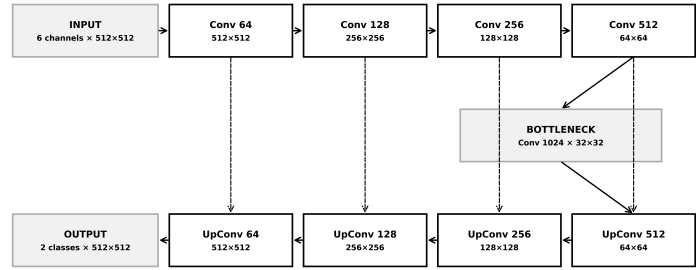}
    \caption{U-Net architecture}
    \label{fig:unet_architecture}
\end{figure}

\noindent
\textbf{DeepLabV3 (Fine-tuned)}\\
We employed a ResNet-50 backbone pretrained on the \textit{ImageNet-1K} dataset and modified its first convolutional layer (\texttt{conv1}) to process as input spectral images of $512 \times 512$ pixels with 6 spectral bands. To preserve the representational capacity of the pretrained weights, we used a mixed initialization strategy as reflected in Equation \ref{eq:pretrained}.
\begin{equation}
W_{new}[i] =
\begin{cases}
W_{pretrained}[i] & \text{for } i \in {0, 1, 2} \text{ (RGB)} \\
\text{mean}(W_{pretrained}) & \text{for } i \in {3, 4, 5} \text{ (Additional Bands)}
\end{cases}
\label{eq:pretrained}
\end{equation}
The classifier head was replaced with a $1 \times 1$ convolution layer outputting two channels corresponding to Water and Background classes.\\

\noindent
\textbf{U-Net (From Scratch)} \\
We constructed a symmetric encoder–decoder U-Net with four downsampling levels. Each level uses two consecutive convolutional layers ($3\times 3$ kernels) followed by Batch Normalization and LeakyReLU activations, facilitating stable gradient propagation from random initialization. Skip connections bridge encoder and decoder blocks by concatenating high-resolution encoder feature maps with the upsampled decoder outputs, enabling the network to retain detailed shoreline structure. Figure \ref{fig:unet_architecture} illustrates a visual overview of this architecture.\\

\noindent
\textbf{DeepWaterMap}\\
The model is a fully convolutional neural network based on the architecture proposed in \cite{article}, specifically designed for large-scale surface water extraction from multispectral satellite imagery. It is computationally efficient, enabling rapid inference over wide geographic areas while maintaining high accuracy and is adapted to the spectral characteristics of water, particularly in the near-infrared (NIR) and shortwave infrared (SWIR) bands, where water exhibits strong absorption. Furthermore, the model uses spectral-aware convolutional layers to improve the discrimination between water and non-water surfaces in difficult scenarios, such as urban environments, shadowed regions, and turbid waters.

\subsection{Training methodology}
To isolate architectural effects, both models were trained under an identical experimental setup. Training was conducted on $512 \times 512$ image patches. To address the strong class imbalance of water pixels, we employed a Water-Biased Sampling strategy that enforces a minimum amount of water in each training patch. All experiments used fixed random seeds to ensure reproducibility.

We adopted a composite loss function that jointly optimizes per-pixel accuracy and regional overlap. In this setup, a dedicated class-weighting scheme was applied to the Cross-Entropy term to penalize misclassification of the minority water class more heavily. A full description of the training environment, including patch size, optimizer configuration, and the chosen set of hyperparameters, is provided in Table \ref{tab:training_params}. All experiments were conducted on the Kaggle cloud-computing platform, using an NVIDIA Tesla P100 GPU with 16 GB of VRAM to speed up the training process.
\begin{table}[h]
\centering
\caption{Model Training Hyperparameters and Optimization Settings}
\label{tab:training_params}
\begin{tabular}{@{}ll@{}}
\toprule
\textbf{Parameter}            & \textbf{Value / Setting} \\ \midrule
Patch Size                    & $512 \times 512$ pixels  \\
Sampling Strategy             & Water-Biased ($\ge 1.5\%$ coverage) \\
Reproducibility               & Fixed Random Seeds       \\
Optimizer                     & AdamW                    \\
Learning Rate (Initial)       & $5 \times 10^{-5}$       \\
Weight Decay                  & $10^{-4}$                \\
Loss Function                 & $\mathcal{L}_{\text{Total}} = 0.5\,\mathcal{L}_{\text{CE}} + 0.5\,\mathcal{L}_{\text{Dice}}$ \\
Class Weights (Land:Water)    & 1:20                     \\
Scheduler                     & Cosine Annealing Warmup  \\
Warmup Period                 & 5 Epochs                 \\
Batch Size                    & 4                        \\
\bottomrule
\end{tabular}
\end{table}
\subsection{Water indices and quality parameters}

This study's importance goes beyond identifying inland water bodies. The high-precision masks from DeepLabV3 and U-Net enable for a diagnostic assessment of the water's composition, using a series of standard water quality indices. For the estimation of these indices, we use six Sentinel-2 bands, namely B02, B03, B04, B08, B11, and B12, which enable us to extract data for turbidity, biological activity, and chemical contamination. 

The NDWI \cite{10765764} serves as a primary metric for open-water detection, relying on the strong reflectance of green wavelength (B03) from vegetation and the strong absorption of water in the Near-Infrared (NIR) band B08.

A variant of the NDWI is the MNDWI \cite{6723425}, which substitutes the NIR band with the Short-Wave Infrared (SWIR) band B11 to more effectively attenuate contributions from urban and built-up surfaces adjacent to water bodies.

Turbidity acts as a proxy for the concentration of Total Suspended Solids (TSS) \cite{7813877}. Elevated turbidity values indicate greater amounts of silt, clay, or anthropogenic pollutants that enhance light scattering in the visible domain. The turbidity index, as given in Table \ref{tab:water_indices}, is based only on the RGB color bands. B02 (blue), B03 (green), B04 (red).
\begin{table}[h]
\centering
\renewcommand{\arraystretch}{1.5}
\caption{Spectral indices for water applications}
\begin{tabular}{llc}
\hline
\textbf{Type of index} & \textbf{Name} & \textbf{Formula} \\
\hline

\multirow{2}{*}{Water presence} 
& NDWI 
& $\frac{B03 - B08}{B03 + B08}$ \\

& MNDWI 
& $\frac{B03 - B11}{B03 + B11}$ \\
\hline

\multirow{3}{*}{\makecell[l]{Water quality\\assessment}} 
& Turbidity 
& $\frac{B04 + B03}{B02}$ \\

& NDCI (algae proxy) 
& $\frac{B03 - B04}{B03 + B04}$ \\

& NDOSI 
& $\frac{B11 - (B02 + B03 + B04)}{B11 + (B02 + B03 + B04)}$ \\
\hline

Spectral depth 
& Relative Bathymetry 
& $1 - \frac{B08}{B08_{max}}$ \\
\hline
\end{tabular}
\label{tab:water_indices}
\end{table}

Algal Biomass \cite{7563846} is inferred by examining the reflectance peak in the green band B03 relative to the absorption in the red band B04. This index is essential for tracking eutrophication. Elevated values can indicate the onset of algal blooms that draw down dissolved oxygen and may trigger hypoxic episodes detrimental to local fish communities.

To identify chemical contaminants such as surface oil, we apply the Normalized Difference Oil Spill Index (NDOSI) \cite{9955062}, which captures the characteristic SWIR response of hydrocarbons.

In order to perform some depth dependent water quality analysis we used the Relative Bathymetry (Spectral Depth Proxy) \cite{9810335}. 
Although not directly yielding absolute depth in meters without field calibration, this a spectral depth metric which exploits the rapid attenuation of NIR radiation, absorbed within the upper centimeters of the water column.

Table \ref{tab:water_indices} provides the formulas for each of the presented indices.

To evaluate the spatial patterns and internal coherence of the water quality indicators, we calculate the local standard deviation ($\sigma$) for the complete set of derived metrics. This statistic allows us to differentiate spatially homogeneous, stable water bodies from regions that display localized spectral anomalies, such as sediment plumes, algal blooms, or surface-active chemical films.

The local spatial variance is determined by moving a $5 \times 5$ window over the masked water pixels. For each spectral index $k$, the local deviation is given by Equation \ref{eq:var}.
\begin{equation}
\sigma_k = \sqrt{\frac{1}{N} \sum_{i=1}^{N} (x_{i,k} - \bar{x}_k)^2}
\label{eq:var}
\end{equation}
where $x_{i,k}$ represents the value of the $i$-th pixel for metric $k$ within the window, $\bar{x}_k$ is the corresponding local arithmetic mean, and $N$ denotes the number of valid water pixels present in the neighborhood.

\subsection{Visualization and color palettes}
Large quantities of numerical results obtained over water surfaces are often difficult to interpret directly. Visual inspection is therefore significantly facilitated by representing these results as heatmaps, which are readily interpretable by human observers and allow for rapid identification of spatial patterns and anomalies.

For a comprehensive qualitative environmental analysis, we generate a suite of metric maps alongside their corresponding variance heatmaps. An important aspect of effective visualization is the correct selection and consistent application of color palettes, which directly impact interpretability and comparability across different metrics.\\

\noindent
\textbf{Color Palette Strategy}\\
Our primary objective in selecting color palettes was to ensure intuitive interpretation and, where appropriate, facilitate direct comparison between different water quality parameters. While a single universal palette for all measurements would offer maximal comparability, the diverse nature and physical implications of each index necessitate a tailored approach to maintain clarity and avoid misinterpretation. Therefore, we adopted a strategy that balances consistency with thematic relevance, specifically:
\begin{itemize}
\item For indices representing similar physical phenomena or requiring direct comparative analysis (e.g., different water presence indices), a consistent color scheme or variations thereof are employed.
\item For indices representing distinct physical properties (e.g., water presence vs. pollution), thematic palettes are chosen to align with common visual associations and scientific conventions, thereby enhancing immediate understanding.
\end{itemize}

\begin{figure}
    \centering
    \includegraphics[width=0.55\linewidth]{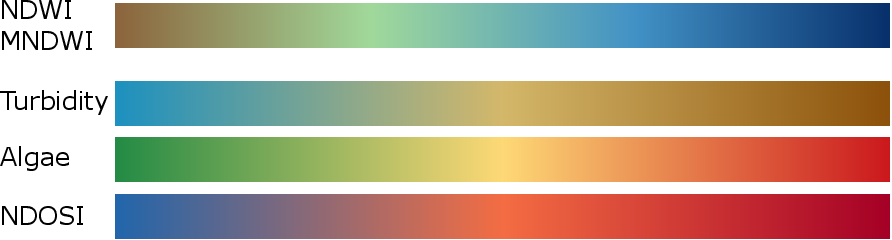}
    \caption{Color palettes for the water spectral indices.}
    \label{fig:palettes}
\end{figure}

\noindent
\textbf{Thematic Color Schemes and Interpretation}\\
The physical state of the water body is illustrated using the following thematic color schemes, each chosen to reflect the specific characteristics of the measured parameter. All generated figures include the corresponding color scales to ensure precise quantitative interpretation of the visual data. The color palettes are represented in Fig. \ref{fig:palettes}.

\noindent
\textbf{Water presence indices}\\
\textit{NDWI/MNDWI:} A continuous brown-to-deep-blue gradient is employed across the $[-1, 1]$ index range. In this scheme, negative values (land and vegetation) are represented in brown-gray, while positive values transition from light cyan-green to deep blue to indicate increasing water presence and clarity. This choice aligns with the common perception of water color and effectively highlights the extent and optical properties of water.\\

\noindent
\textbf{Water quality indices}\\
\textit{Turbidity:} A blue-to-brown sequential scale is utilized. Clear, low-turbidity waters are represented by blue, gradually shifting to brown as the concentration of suspended particulate matter increases. This progression intuitively reflects the visual appearance of water becoming murkier. Specific thresholds for turbidity (e.g., blue for $< 5$ NTU, light brown for 5-25 NTU, dark brown for $> 25$ NTU) are considered to align with established environmental classifications.\\

\noindent
\textit{Algae (Chlorophyll-a):} A green-to-red diverging scale is used to highlight phytoplankton biomass. Green indicates healthy, lower algal presence, while a shift towards red signifies elevated chlorophyll-a concentrations, indicative of potential eutrophication and algal blooms. This palette allows for quick identification of areas requiring attention. Typical ranges (e.g., green for $< 10$ \(\mu g/L\), yellow for 10-50 \(\mu g/L\), red for $> 50$ \(\mu g/L\)) are used for interpretation.\\

\noindent
\textit{NDOSI/Oil Ratio:} A blue-to-dark-red alert scheme is specifically designed to detect hydrocarbon pollution. Blue signifies no or negligible oil presence, while increasing saturation towards dark red indicates higher probabilities or concentrations of oil slicks, serving as a clear warning signal.\\

\noindent
\textbf{Spectral Depth}\\
A light-to-dark blue monochromatic scale indicates relative bathymetry. Lighter shades represent shallower areas, progressively deepening to darker blues for greater depths. This provides an intuitive visual cue for the underwater topography.\\

\noindent
\textbf{Homogeneity Heatmaps (Variance)}\\
Spatial heterogeneity across all indices is depicted using a blue-to-red sequential gradient. A blue signal corresponds to highly stable and uniform water conditions (\(\sigma < 0.15\)), whereas a red signal denotes pronounced local variability (\(\sigma > 0.35\)). Intermediate values transition through green and yellow. This gradient effectively visualizes areas of energetic mixing zones, intricate land–water transition areas, or localized disturbances.

\section{Results and Discussions}
This section is dedicated to assessing the efficacy of deep learning and spectral techniques for water body segmentation, followed by a spatial analysis of water quality indicators. Initially, a robust segmentation baseline will be established. Subsequently, we will investigate the distribution of turbidity, algal presence, and mineral signatures within Lake Dumbrăvița, a representative study site for the Brașov region.

\subsection{Water body segmentation}
The first step in monitoring inland water bodies and assessing water quality, is the accurate segmentation of these surfaces in the targeted satellite images, in our case Sentinel-2 multispectral data. Therefore, we trained two CNN-based models, DeepLabV3 and a custom U-Net, and evaluated their performance against the state-of-the-art DeepWaterMap model and conventional spectral baselines. These comparisons employed both the continuous NDWI and MNDWI results as well as their thresholded binary masks.

The results obtained for both trained models are promising, with the U-Net architecture performing slightly better than the pretrained DeepLabV3 and outperforming the DeepWaterMap benchmark with higher recall and specificity, resulting in a more dependable mapping of intricate water boundaries. Although DeepWaterMap achieves strong accuracy at the global scale, it is comparatively less responsive to the finer local characteristics present in the Sentinel-2 imagery used here. By contrast, the U-Net architecture more effectively captures detailed shoreline structures and smaller water bodies. Table \ref{table:res} presents a quantitative comparison of the models, reporting the validation metrics for each model at the epochs where their performance was most stable.
\begin{table}[h]
\centering
\caption{Validation Results Summary}\label{table:res}
\begin{tabular}{lllr}
\toprule
\textbf{Metric} & \textbf{DeeplabV3} & \textbf{U-Net} & \textbf{DeepWaterMap}\\
\midrule
Accuracy & 96.31\% & 97.38\% & 98.18\%\\
IoU & 81.09\% & 87.47\% & 71.84\%\\
F1-Score (Dice) & 86.88\% & 92.24\% & 77.34\%\\
Recall & 89.39\% & 94.71\% & 77.28\%\\
Precision & 85.95\% & 91.83\% & 82.63\%\\
Specificity & 75.19\% & 96.29\% & 85.63\%\\
\bottomrule
\end{tabular}
\end{table}
As reported in its original publication \cite{article}, the DeepWaterMap model reported class-specific scores of 88\% for both Water precision and Water recall, considered on a dataset composed of Landsat images. In contrast, when evaluated on the Sentinel-2 dataset used in this study with 25 patches per image, the pre-trained DeepWaterMap model attained the precision and recall metrics summarized in Table \ref{table:res}. It should be noted that the spectral bands of our Sentinel-2 dataset correspond to those used in the Landsat input images for DeepWaterMap, ensuring model compatibility.% between the two models.

The comparative analysis indicates that, although all three models can reliably detect surface water, the custom U-Net architecture delivers the most robust performance for multispectral water segmentation. In quantitative terms, U-Net achieved the highest values across almost all key metrics. Its markedly higher specificity, relative to DeepLabV3, highlights a stronger capacity to separate water bodies from complex terrestrial environments, thereby reducing false-positive detections. The only metric in which DeepWaterMap slightly outperforms the U-Net model is accuracy. However, the difference (less than $1\%$) is not significant for the overall results. The results indicate that training directly on this dataset yields improved quantitative metrics as well as richer qualitative detail.

\begin{figure}[h]
    \centering
    \includegraphics[width=0.35\textwidth]{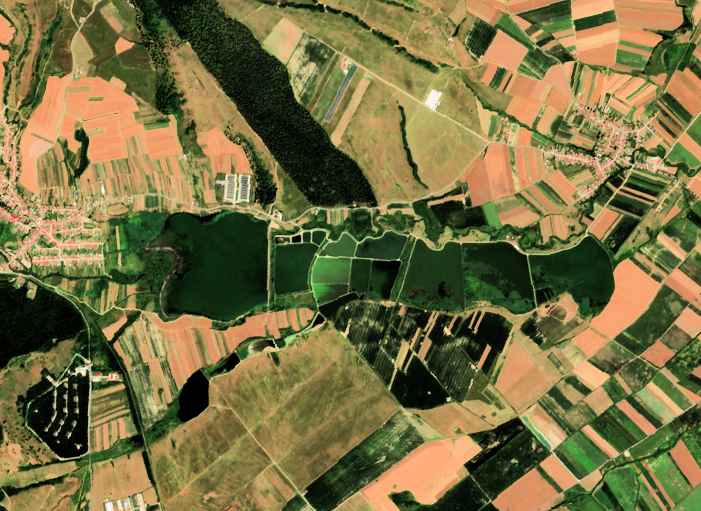}
    \includegraphics[width=0.35\textwidth]{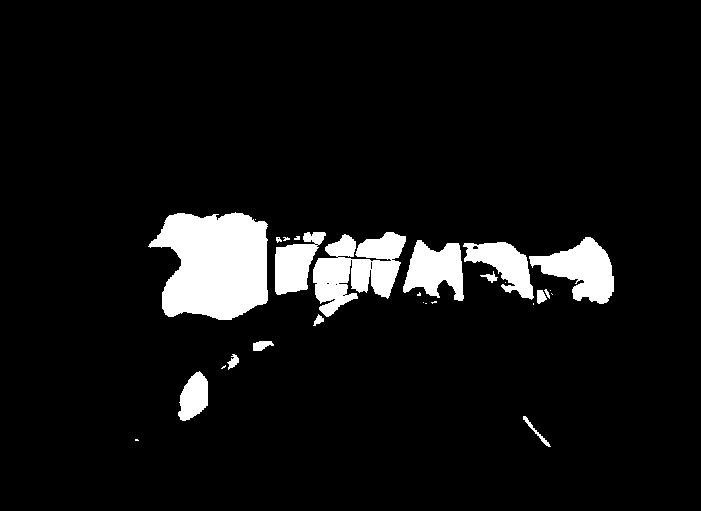}
    \includegraphics[width=0.35\textwidth]{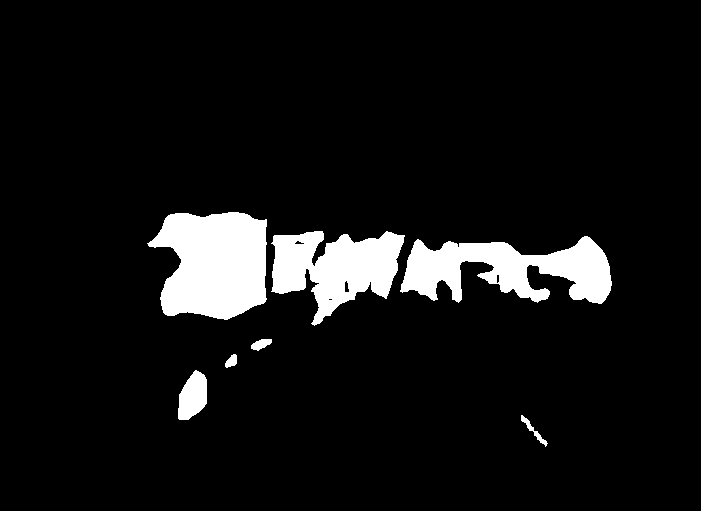}
    \includegraphics[width=0.35\textwidth]{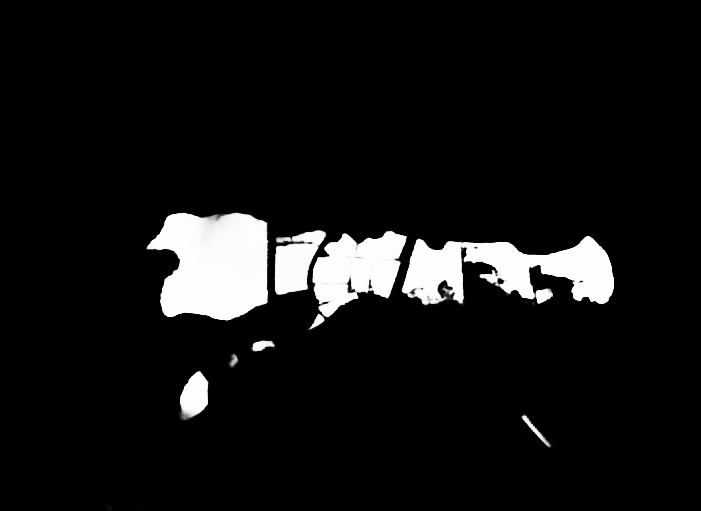}
    \caption{Original image (B02, B03, B04 bands); U-Net result; DeepLabV3 result; DeepWaterMap result.}
    \label{fig:visual_comparison}
\end{figure}

\begin{figure}[h]
    \centering
    \includegraphics[width=0.40\textwidth]{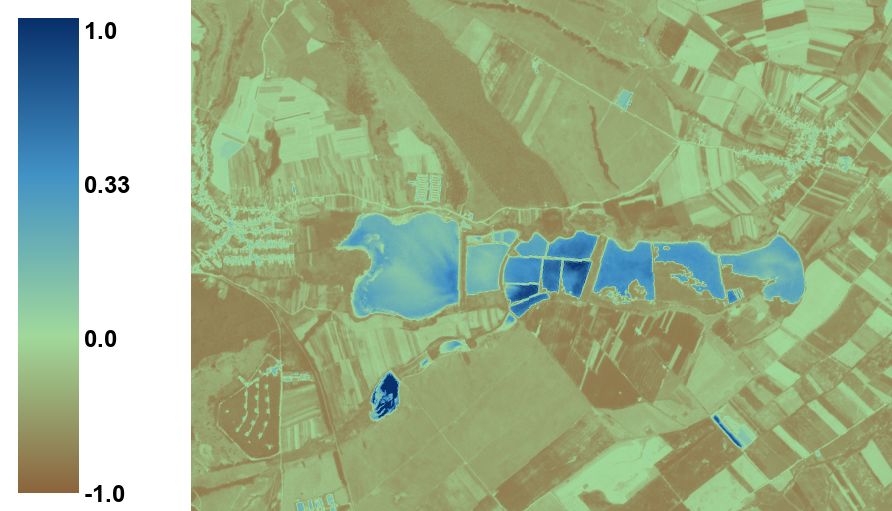}
    \includegraphics[width=0.40\textwidth]{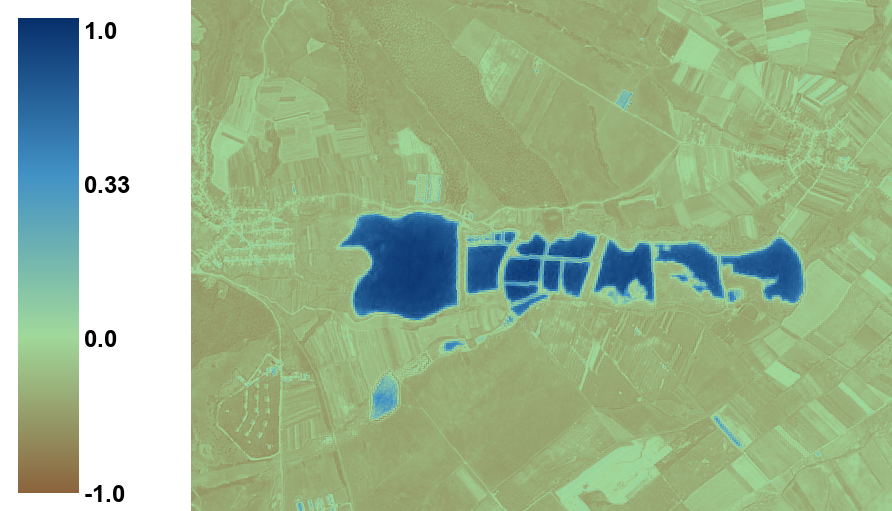}
    %\caption{Visual representation of NDWI (left) and MNDWI (right) color index maps}
    %\label{fig:ndwi_mndwi}
%\end{figure}

%\begin{figure}[h]
%    \centering
    \includegraphics[width=0.37\textwidth]{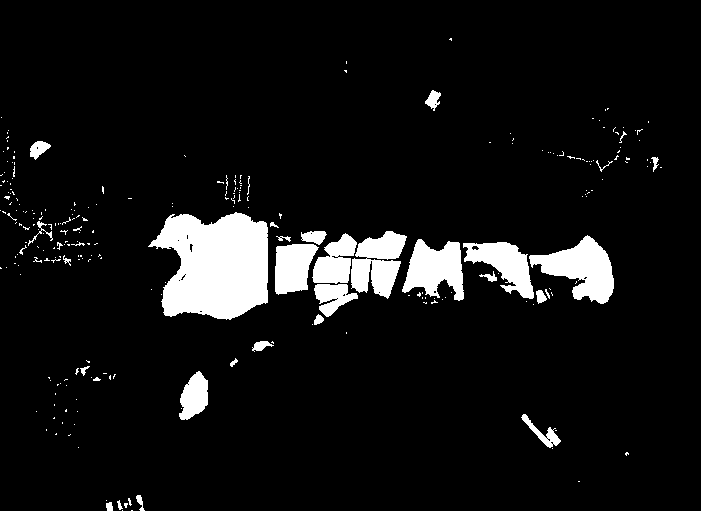}
    \includegraphics[width=0.37\textwidth]{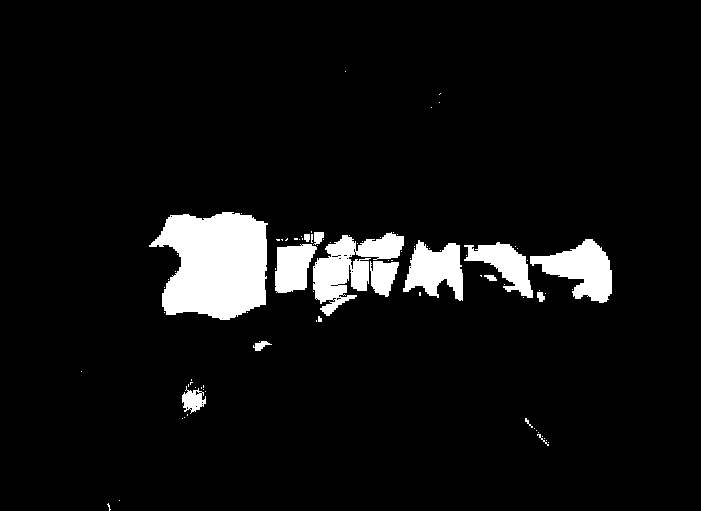}
    \includegraphics[width=0.37\textwidth]{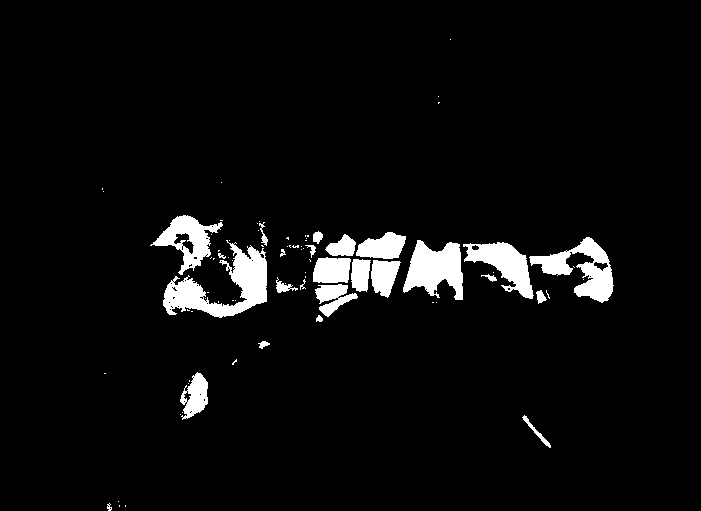}
    \includegraphics[width=0.37\textwidth]{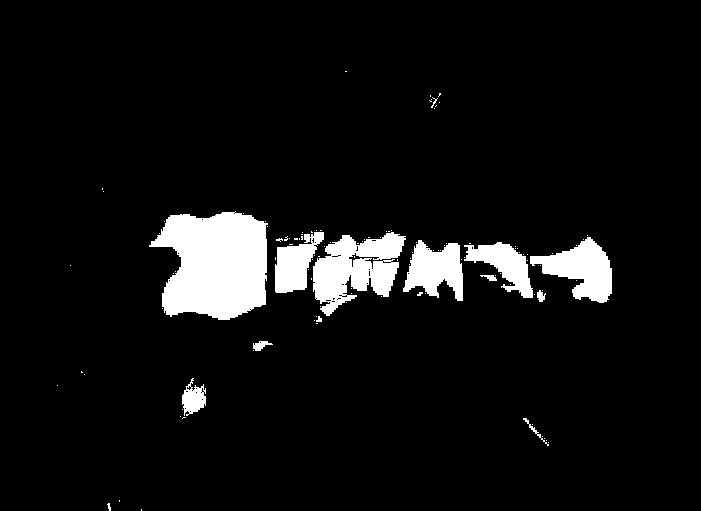}
   
    \caption{Visual representation of NDWI (top-left) and MNDWI (top-right) color index maps and binary water masks from NDWI (left) and MNDWI (right); middle row: Otsu's automatic threshold; bottom row: fixed threshold of 0.0.} 
    \label{fig:ndwi_mndwi_binary}
\end{figure}

The visual comparison of the segmentation results, as presented Fig. \ref{fig:visual_comparison}, demonstrates that U-Net produces a considerably more continuous and finely detailed water mask that accurately traces the irregular and highly detailed lake shorelines, thereby preserving spatial consistency. Moreover, the mask produced by the U-Net successfully avoids classifying the narrow non-water bands corresponding to the dams separating the different pools of Lake Dumbrăvița as water, whereas in the other models only the main dams are distinguished, with DeepWaterMap achieving slightly better segmentation than DeepLabV3. By contrast, the DeepWaterMap predictions display visible fragmentation and scattered noise artifacts in the surrounding areas.
To provide a visual baseline for water body detection, Fig. \ref{fig:ndwi_mndwi_binary} top row illustrates the NDWI and MNDWI indices computed directly from the original multispectral data prior to the segmentation process. To assess the performance of the NDWI and MNDWI indices, binary water masks were generated using both automated and fixed thresholding methods, as shown in Fig. \ref{fig:ndwi_mndwi_binary}.

An automated strategy based on Otsu’s method \cite{rs8040354} was evaluated to provide a comparative reference. 
This assessment highlights the fundamental drawbacks of spectral thresholding. Specifically, Otsu’s algorithm often produces false positives by enforcing a bimodal partition even in areas without water, whereas fixed thresholds are highly vulnerable to mixed-pixel effects and changes in water properties. Such limitations can lead to noisy background, by false positive pixel detection, or inaccuracy through false negative, when water pixels fall beyond the automatic Otsu threshold. The results of Otsu threshold on the NDWI and MNDWI maps for Lake Dumbrăvița are presented in Fig. \ref{fig:ndwi_mndwi_binary}.

Although a global threshold of 0.0 on the NDWI, respectively the MDNWI, is frequently cited as a standard baseline \cite{su13179798}, the visual results in Fig.~\ref{fig:ndwi_mndwi_binary} show significantly less water being detected, especially in the center of the water body. 
Even more restrictive threshold recommendations from the literature, such as 0.2 for NDWI \cite{eos_ndwi} and 0.35 for MNDWI \cite{article3}, yielded even poorer results on the selected Lake Dumbrăvița example. These inconsistencies show that using fixed spectral thresholds involves a delicate balance between reducing noise and accurately detecting features. Deep learning models address this by using learned spatial context instead of simple intensity cutoffs.

\subsection{Estimation of water quality parameters}
\begin{figure}[h]
\centering
\includegraphics[width=0.40\textwidth]{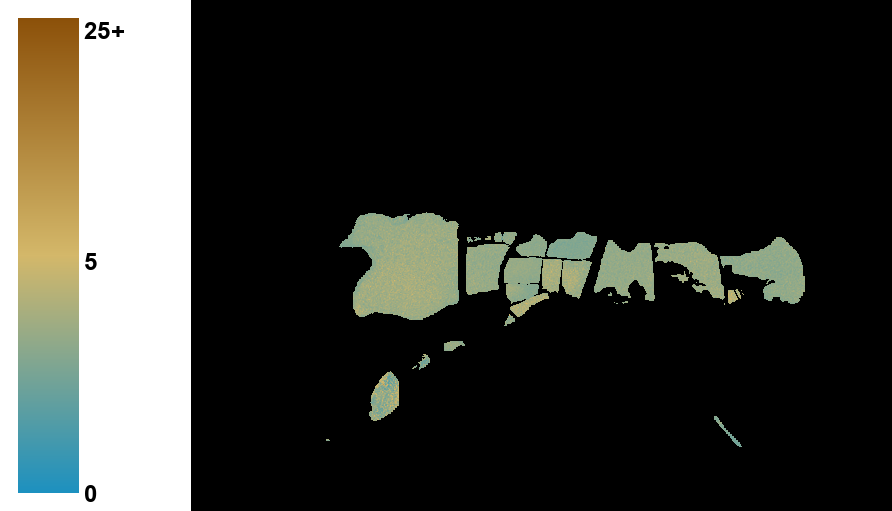}
\includegraphics[width=0.40\textwidth]{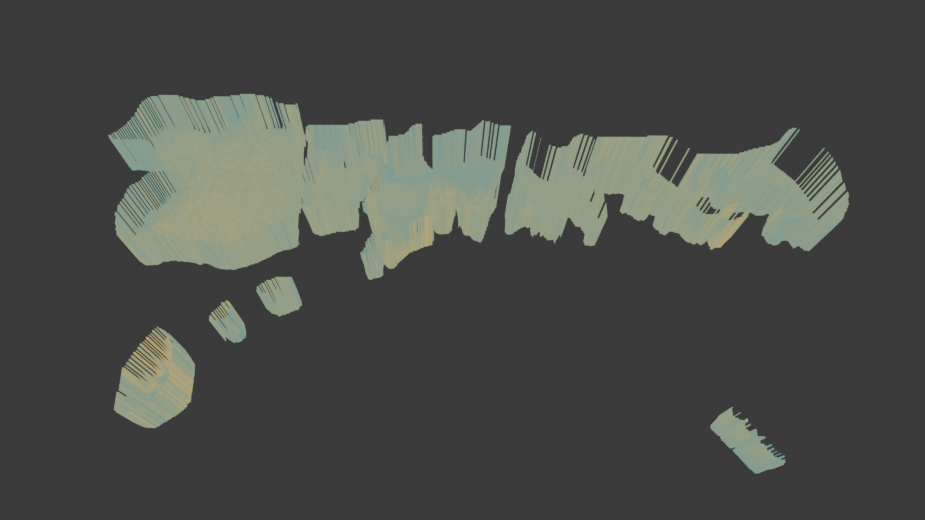}
\includegraphics[width=0.40\textwidth]{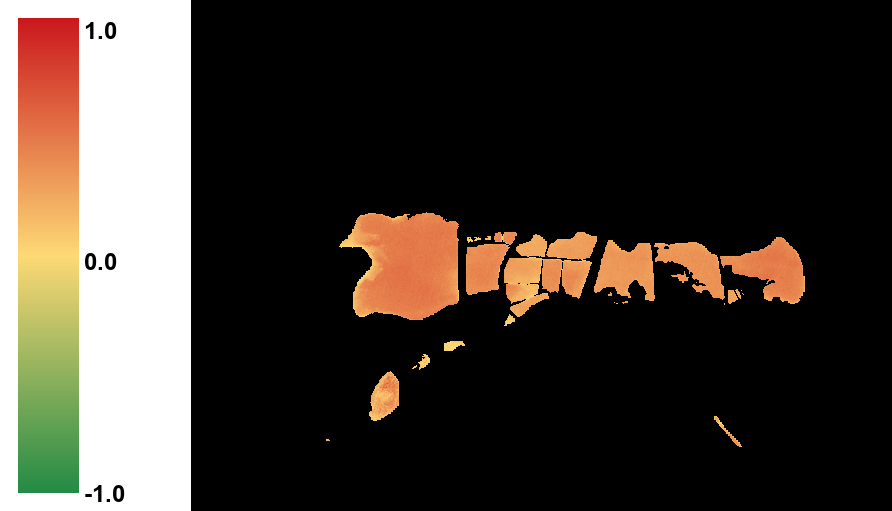}
\includegraphics[width=0.40\textwidth]{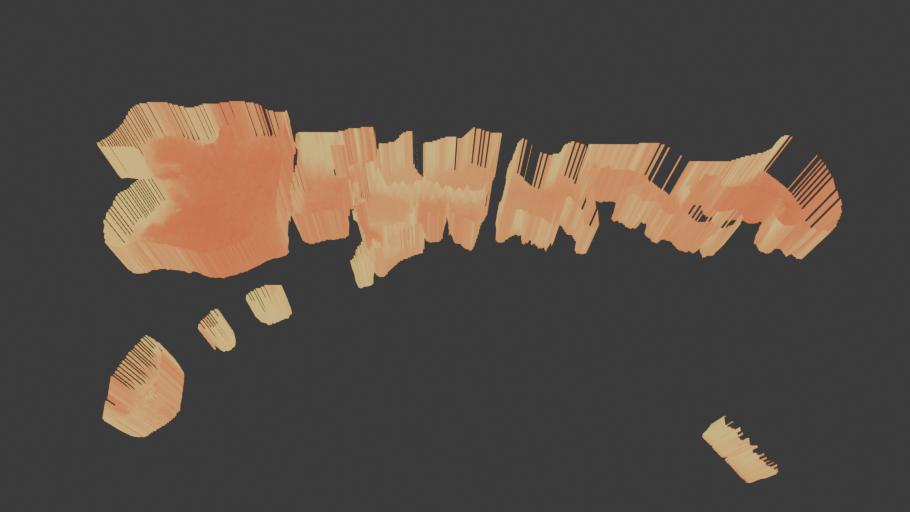}
\includegraphics[width=0.40\textwidth]{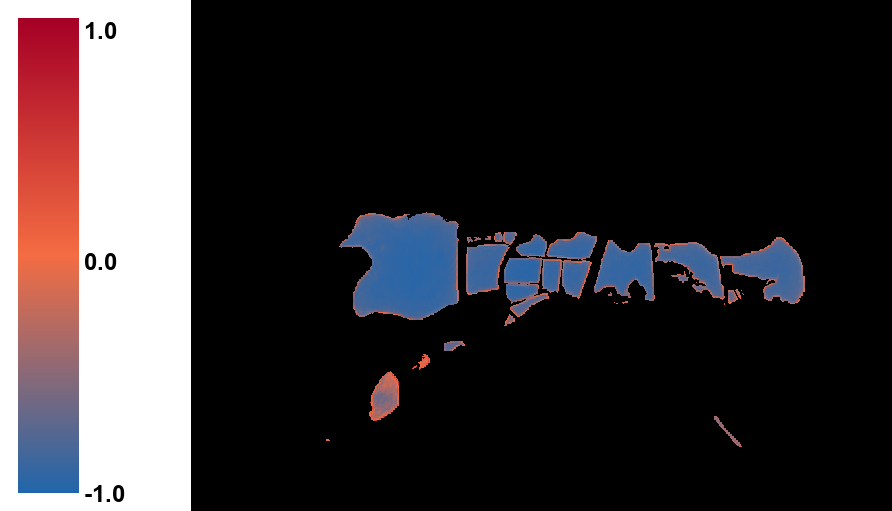}
\includegraphics[width=0.40\textwidth]{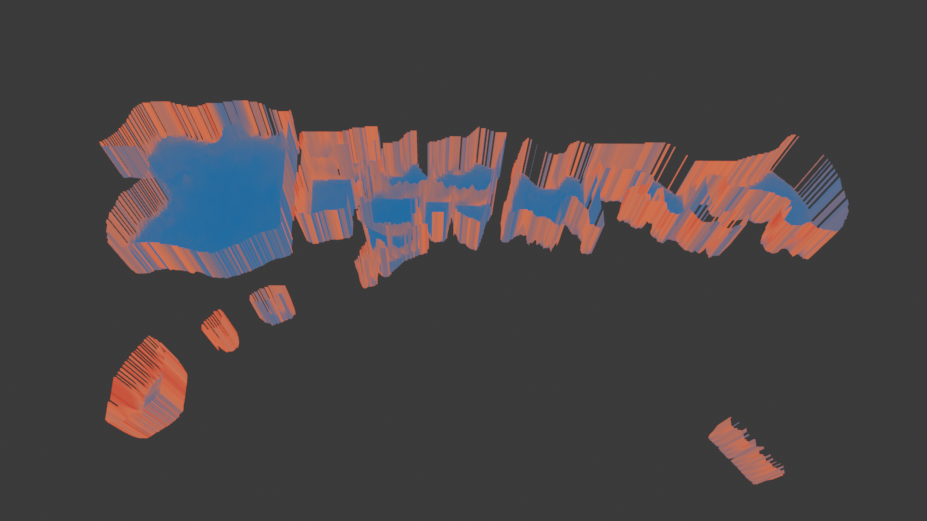}
\caption{Spatial analysis of Lake Dumbrăvița: Turbidity (top row), Algae index (middle row), and NDOSI (bottom row).}
\label{fig:metrics_visual}
\end{figure}

Using the water masks generated by the U-Net model, we carried out a diagnostic assessment of Lake Dumbrăvița’s water quality, see Fig. \ref{fig:metrics_visual}. For every metric, the picture offers a multi-perspective view. Rows show the numerical colour scale, the 2D spatial map, and the 3D bathymetric visualisation from left to right.

The turbidity maps reveal a largely uniform spatial distribution across the basins, with values predominantly falling within the $3.39 \pm 0.37$ range. This consistency suggests a widespread presence of suspended sediment loads that restrict light penetration and reduce photosynthetic activity throughout these low-energy areas. Similarly, algal concentrations appear uniformly distributed throughout the basins, as reflected by consistently high values $(>0.25)$. This even spatial pattern suggests that the environmental factors controlling algal development remain fairly constant across the whole area. This increased algal biomass approaches the 0.30 threshold associated with eutrophication and the potential onset of harmful algal blooms, posing ecological risks such as oxygen depletion and disruption of food-web structure. The NDOSI index highlights a linear feature $(>0.20)$ along the southern shoreline, corresponding to zones of fine-sediment accumulation driven by preferential alongshore transport.

\begin{figure}[h]
\centering
\includegraphics[width=0.28\textwidth]{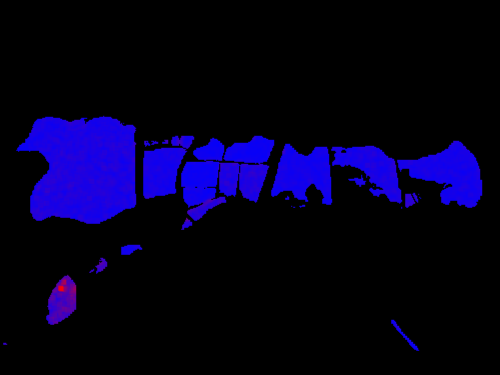}
\includegraphics[width=0.28\textwidth]{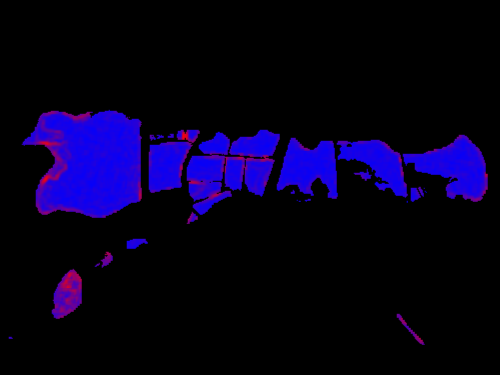}
\includegraphics[width=0.28\textwidth]{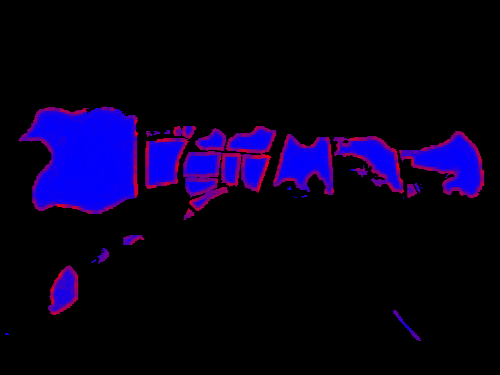}
\caption{The corresponding local deviation (variance) maps for each metric: Turbidity (left), Algae (middle), and NDOSI (right).}
\label{fig:variance_visualisation}
\end{figure}

To assess the robustness of these spectral metrics, we analyzed their local variability, see Fig. \ref{fig:variance_visualisation}. While shorelines show the highest variability ($\pm$30 - 50\% standard deviation) due to mixed pixels, atmospheric correction issues, and shallow-water reflections, the open water areas are very stable ($\pm$10 - 20\% variance). This means about 70\% of the lake's surface,  specifically  areas more than 100 m from the shore, provides reliable data for quantitative analysis in these consistent, deeper waters.

\begin{figure}[h]
    \centering
    \subfloat[]{\includegraphics[width=0.40\textwidth]{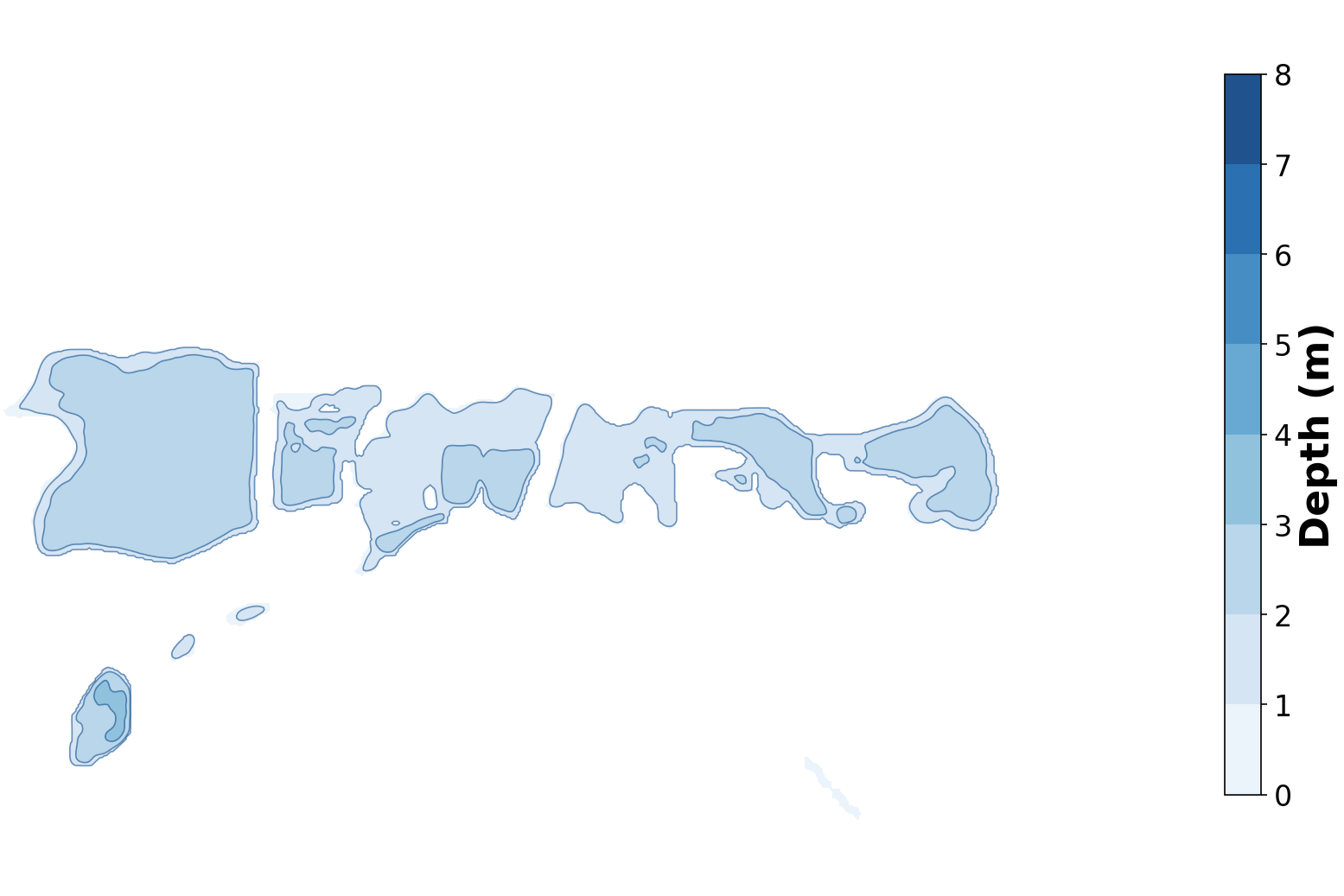}
    \label{fig:depth_map}}
    \hfill
    \subfloat[]{\includegraphics[width=0.40\textwidth]{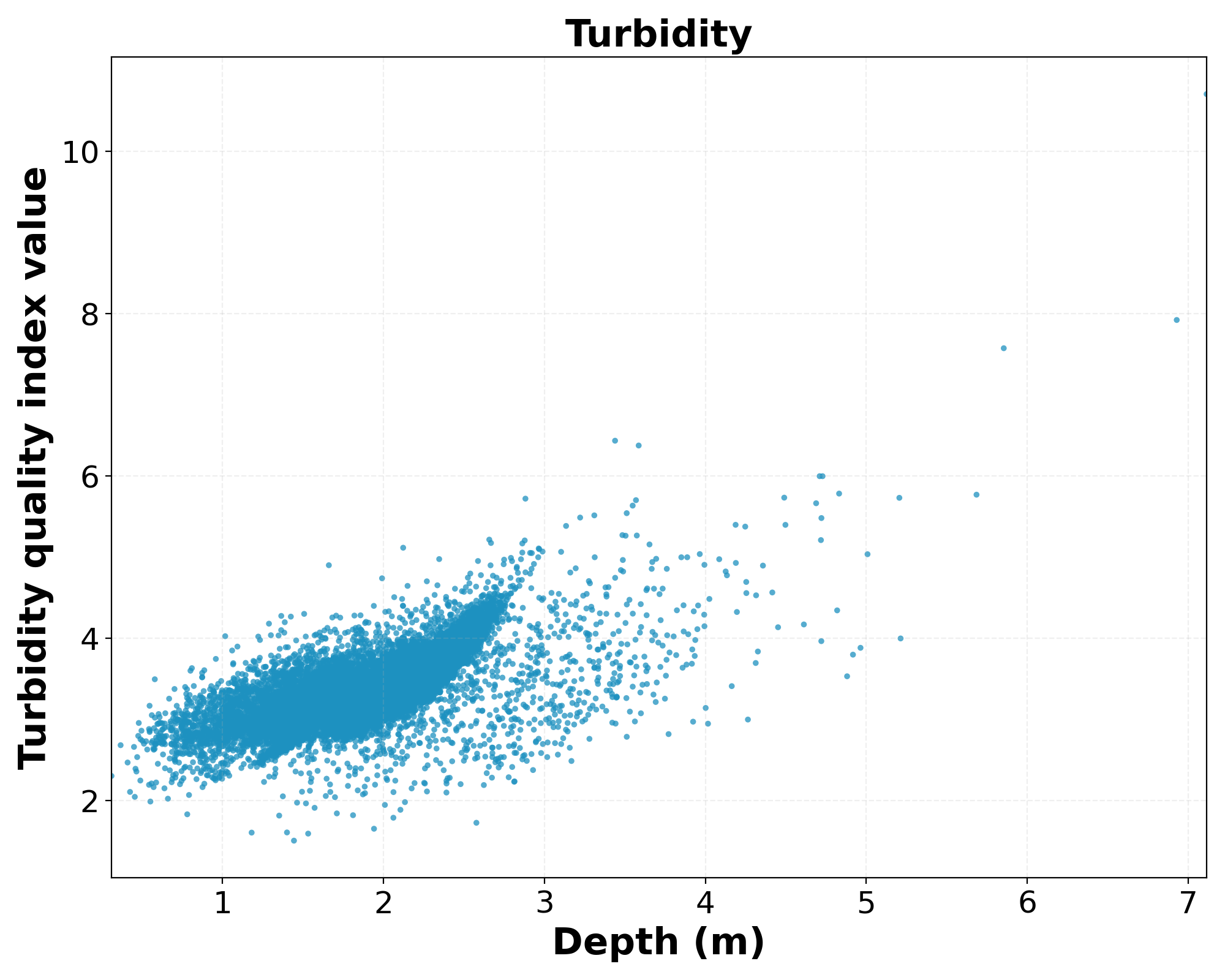}
    \label{fig:turbidity_depth}}
    
    \subfloat[]{\includegraphics[width=0.40\textwidth]{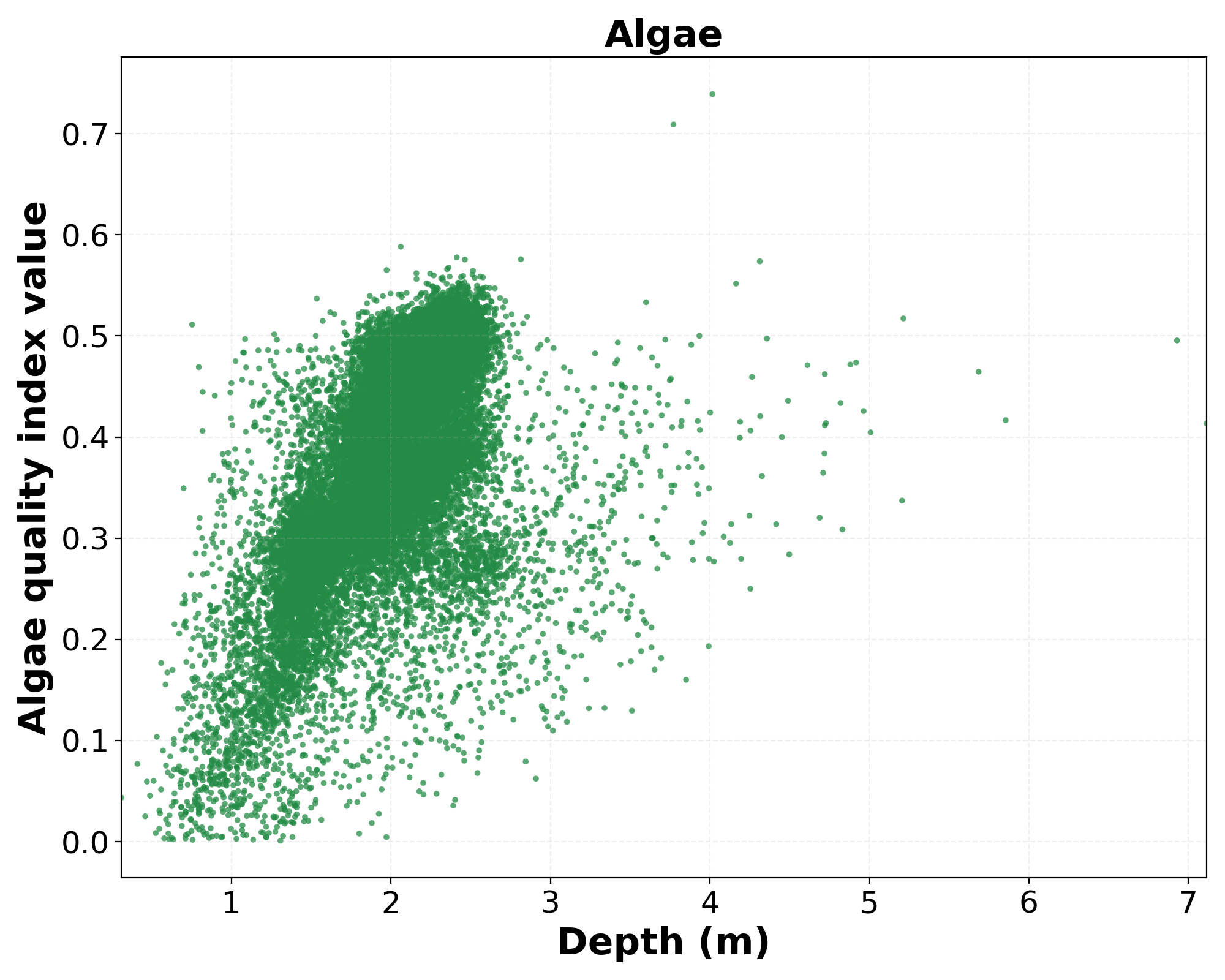}
    \label{fig:algae_depth}}
    \hfill
    \subfloat[]{\includegraphics[width=0.40\textwidth]{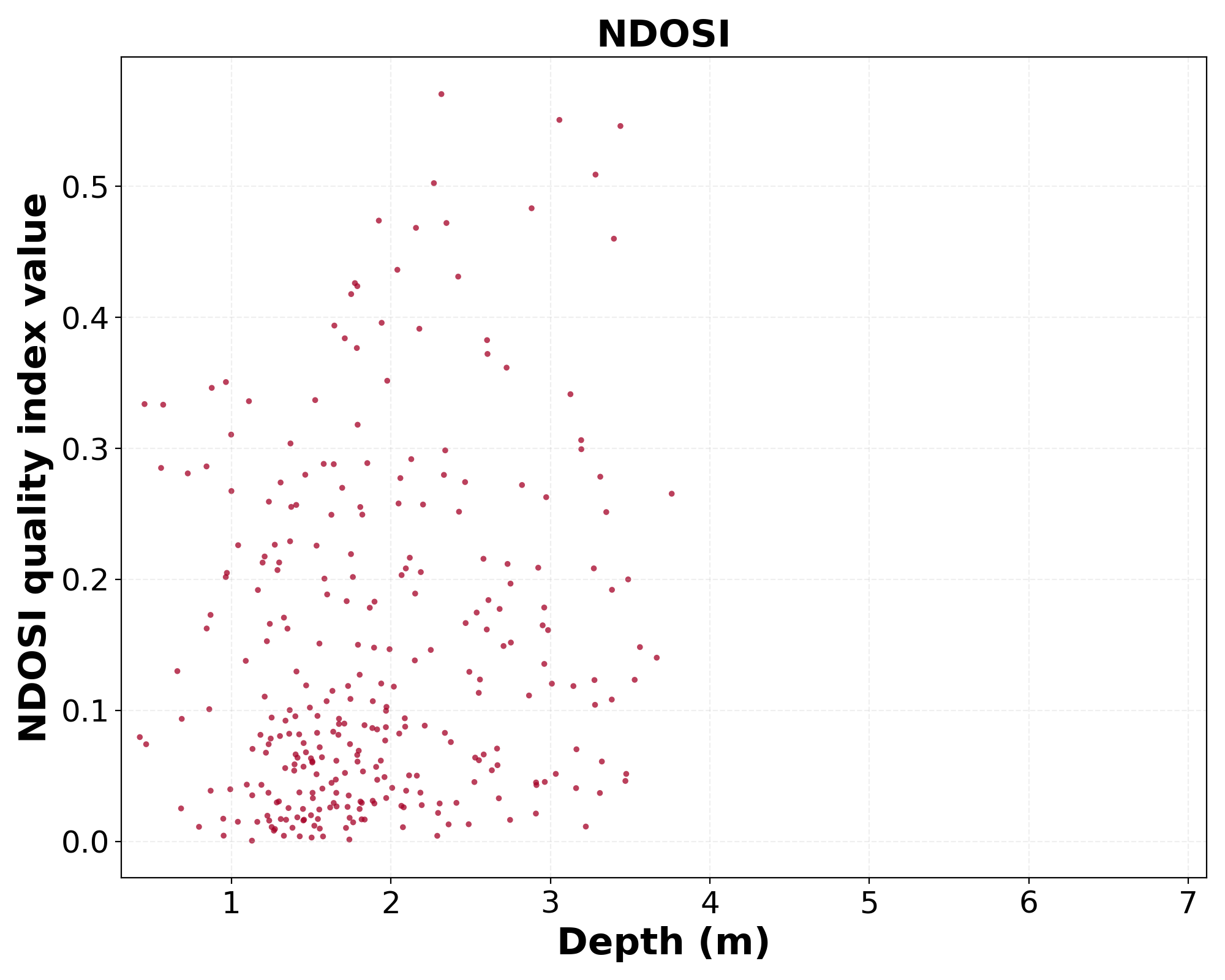}
    \label{fig:ndosi_depth}}
       
    \caption{Spatial and per-pixel distributions of water quality indices relative to water depth: (a) estimated bathymetric map; (b) Turbidity vs. depth; (c) Algae vs. depth; and (d) NDOSI vs. depth.} 
    \label{fig:depth_and_indices}
\end{figure}

The bathymetric pattern illustrated in Fig. \ref{fig:depth_map} provides a relative depth estimate derived from spectral reflectance, which we use as a proxy in lieu of absolute in-situ calibration data. Although the absence of local ground-truth measurements introduces uncertainty in the absolute depth values, this approach still enables us to analyze the interrelationships among the water-related metrics presented in Fig. \ref{fig:metrics_visual}.

Plotting index values against estimated depth reveals a clear bathymetric dependency. Turbidity (Fig. \ref{fig:turbidity_depth}) shows a non-linear, two-phase evolution: stable in shallow areas, then increasing significantly with depth past 2.0 m, peaking at the reservoir's maximum depth. Algal signatures (Fig. \ref{fig:algae_depth}) cluster densely between 1.5 - 2.5 m depth, suggesting that while biological activity is widespread, rising turbidity at greater depths likely limits light and thus biomass. In contrast, NDOSI (Fig. \ref{fig:ndosi_depth}) values remain low ($<0.5$) and scattered across 0.5 - 3.5 m. These findings indicate Lake Dumbrăvița is a bathymetry-controlled system, where water depth is a controlling factor for sediment resuspension and primary production distribution.

\section{Conclusions and future work}

In this paper, we investigated the most effective approach for inland water body segmentation by comparing three deep learning models: a pretrained DeepLabV3, a custom U-Net trained from scratch, and the state-of-the-art DeepWaterMap. We further benchmarked their outputs against classical NDWI and MNDWI thresholding using a Sentinel-2 image of Lake Dumbrăvița. Based on quantitative performance metrics and visual inspection, we found that the custom U-Net configuration achieved the most accurate and robust results, while all deep learning methods clearly outperformed traditional index-based thresholding.

Water segmentation represents only a first step toward comprehensive water quality monitoring. Spectral water quality indices are essential in this context, and their usefulness is correlated with a meaningful visual representation. Current practices suffer from significant limitations, including perceptual distortions introduced by non-uniform colormaps (e.g., Jet or Rainbow), regional inconsistencies due to the lack of standardized color conventions for water quality, and poor accessibility for color-vision deficient users. To address these issues, we proposed a structured visualization framework based on perceptually uniform and CVD-friendly colormaps, grounded in the Forel–Ule scale as a physically meaningful reference for water color and trophic state. This ensures that spatial patterns in water quality maps remain both scientifically reliable and visually interpretable. Additionally, we performed a spatial correlation analysis between retrieved spectral depth and water quality indicators for Lake Dumbrăvița, demonstrating meaningful relationships through corresponding visual and statistical analyses.

Future work will focus on temporal analysis of water quality indices to assess seasonal and interannual trends, as well as in situ validation to improve calibration and reliability of satellite-derived measurements. We also plan to conduct a broader comparison of machine learning models to identify optimal trade-offs between accuracy, inference time, and model size, enabling real-time deployment on lightweight and edge devices.

\section*{Acknowledgments}
This work was partially funded by the Romanian Ministry of Research, Innovation and Digitalization, SOLUTIONS project entitled ”IMINT for the Black Sea (IMINT)”, contract no. 21Sol(T21)/18.07.2024.

\bibliographystyle{splncs04}
\bibliography{ref}

\end{document}